\documentclass[runningheads]{llncs}
\usepackage[T1]{fontenc}
\usepackage{graphicx}

\usepackage[numbers]{natbib} 

\usepackage{todonotes}
\usepackage{comment}
\usepackage{amssymb}
\usepackage{amsmath}
\usepackage{booktabs}

\usepackage{algorithmic}
\usepackage{array}

\usepackage{stfloats}
\usepackage{times} 

\begin{document}
\title{Understanding the properties and limitations of contrastive learning for Out-of-Distribution detection}%
\titlerunning{Contrastive learning for Out-of-Distribution detection}
%

\author{Nawid Keshtmand\inst{1}\and
Raul Santos-Rodriguez\inst{2}\and
Jonathan Lawry\inst{2}}
\authorrunning{N. Keshtmand, R. Santos-Rodriguez, J. Lawry}

%
\institute{Dept. Aerospace Engineering, University of Bristol\\
\email{yl18410@bristol.ac.uk}\\
\and
Dept. Engineering Mathematics, University of Bristol
\email{\{enrsr,j.lawry\}@bristol.ac.uk}}

\maketitle  

\begin{abstract}
A recent popular approach to out-of-distribution (OOD) detection is based on a self-supervised learning technique referred to as \textit{contrastive learning}. There are two main variants of contrastive learning, namely instance and class discrimination, targeting features that can discriminate between different instances for the former, and different classes for the latter.

In this paper, we aim to understand the effectiveness and limitation of existing contrastive learning methods for OOD detection. We approach this in 3 ways. First, we systematically study the performance difference between the instance discrimination and supervised contrastive learning variants in different OOD detection settings. Second, we study which in-distribution (ID) classes OOD data tend to be classified into. Finally, we study the spectral decay property of the different contrastive learning approaches and examine how it correlates with OOD detection performance. In scenarios where the ID and OOD datasets are sufficiently different from one another, we see that instance discrimination, in the absence of fine-tuning, is competitive with supervised approaches in OOD detection. We see that OOD samples tend to be classified into classes that have a distribution similar to the distribution of the entire dataset. Furthermore, we show that contrastive learning learns a feature space that contains singular vectors containing several directions with a high variance which can be detrimental or beneficial to OOD detection depending on the inference approach used.

\end{abstract}

\section{Introduction}
In recent years, neural networks have been increasingly deployed for prediction tasks.
However, they still make erroneous predictions when exposed to inputs from an unfamiliar distribution \cite{nguyen2015deep}. This poses a significant obstacle to the deployment of neural networks in safety-critical applications. Therefore, for applications in these domains, it is necessary to be able to distinguish in-distribution (ID) data from data belonging to a different distribution on which the neural network was trained (OOD data). 
The problem of detecting such inputs is generally referred to as \textit{out-of-distribution} (OOD) detection, outlier detection, or anomaly detection \cite{ruff2021unifying}.

Recent studies have shown that a type of self-supervised learning referred to as \textit{contrastive learning} can learn a feature space that is able to capture the features of an input $x$ that enable effective OOD detection \cite{chen2020simple,winkens2020contrastive,tack2020csi}. The two main types of contrastive learning approaches are referred to as unsupervised instance discrimination and supervised contrastive learning/supervised class discrimination. The former works by bringing together the features of the different views of the same data point close together in the feature space enabling discrimination between different instances \cite{chen2020simple}. Instead, supervised contrastive learning brings data points belonging to the same class close together in feature space, which enables discrimination between data points belonging to different classes \cite{khosla2020supervised}. In light of the variety of contrastive learning approaches used in practice, it is difficult to ascertain which properties of contrastive learning are beneficial to OOD detection.
In this work, we aim to understand the effectiveness and limitations of existing contrastive learning methods for OOD detection. We approach this in 3 ways.
\begin{enumerate}
    \item We systematically study the performance difference of the instance discrimination and supervised contrastive learning variants in different OOD detection settings (Section \ref{Point_OOD_results}).
    \item We investigate which ID classes OOD data tend to be classified into.
    \item We analyze the spectral decay property of the different contrastive learning approaches and how it correlates with OOD detection performance. The spectral decay indicates how many different directions of high variance are present in the feature space and is related to the feature space compression \cite{roth2020revisiting}.
\end{enumerate}
This involves designing a fair training setup, categorizing the difficulty of OOD detection based on whether ID and OOD datasets are semantically similar (near-OOD) or unrelated (far-OOD), as well as comparing the OOD detection performance using established metrics (Section \ref{Experimental setup}).
 Our main findings and contributions are summarized below:
The key contributions of the paper are as follows:
\begin{itemize}
    \item We perform experiments that show unsupervised instance discrimination, in the absence of fine-tuning, is well-suited to far-OOD detection, where it can be competitive with supervised approaches but ineffective at near-OOD detection (Section \ref{Point_OOD_results}).
    \item We analyze the classification of OOD samples and see that they tend to be classified into classes that have a distribution similar to the distribution of the entire dataset. This suggests that similar to the phenomena seen in deep generative models, discriminative models assign OOD samples to classes where the latent features capture non-class discriminatory information rather than class-discriminatory information \cite{serra2019input} (Section \ref{overall_class_divergence_analysis}).
    \item We show that contrastive learning learns a feature space that contains singular vectors containing several directions with a high variance which can be detrimental or beneficial to OOD detection depending on the inference approach used (Section \ref{variation_factors_analysis}). 
\end{itemize}

\section{Background}
Most OOD detection methods use the features obtained from a classification network trained with the supervised cross-entropy loss. These classification networks learn decision boundaries that discriminate between the classes in the training set. However, using a conventional cross-entropy loss does not incentivize a network to learn features beyond the minimum necessary to discriminate between the different classes and therefore may not be well suited for OOD detection. An alternative to using classification networks for  learning is contrastive learning \cite{chopra2005learning}. The idea of contrastive learning is to learn an encoder $f_\theta$, where $\theta$ denotes the parameters of the encoder, to extract the necessary information to distinguish between similar and other samples. Let $x$ be a query, whilst $\left \{x_{+}  \right \}$ and $\left \{x_{-}  \right \}$ denote the sets of positive samples of $x$ and negative samples of $x$ respectively. The goal of contrastive loss is to learn features of $x$, denoted $z=f_\theta(x)$, similar to the features of $\left \{x_{+} \right \}$ denoted $\left \{z_{+}  \right \} = f_\theta(\left \{x_{+}  \right \})$ whilst also being dissimilar to $\left \{z_{-}  \right \} = f_\theta(\left \{x_{-}  \right \})$. The most common form of similarity measure between features $z$ and $z^\prime$ is given by the cosine similarity as given in Eqn. \ref{eqn: cosine_similarity}: 
\begin{equation}
    sim(z,z') = \frac{z \cdot z'}{\left \| z \right \|\left \| z' \right \|}
    \label{eqn: cosine_similarity}
\end{equation}
The first type of contrastive loss considered in this paper is the Momentum Contrast (Moco) loss \cite{he2020momentum}, where the objective relates to the task of instance discrimination and utilizes two different encoders, a query encoder $f_{\theta_{q}}$ and a key encoder $f_{\theta_{k}}$. The parameters of the query encoder are updated using traditional backpropagation, whilst the parameters of the key encoder are an exponentially weighted average of the parameters of the query encoder, shown by Eqn. \ref{eqn: key_update}: 
\begin{equation}
    \theta_{k}^{'} = m\theta_{k} + (1-m)\theta_{q}
    \label{eqn: key_update}
\end{equation}
where $m$ is an encoder momentum hyperparameter, $\theta_{q}$ are parameters of the query encoder, and $\theta_{k}$ and $ \theta_{k}^{'}$ are the parameters of the key encoder before and after the update respectively.

The setup for the training process is as follows. Let $\widetilde{x}^{1}_{i}$ and $\widetilde{x}^{2}_{i}$ be two independent augmentations of $x_{i}$, namely  $\widetilde{x}^{1}_{i} = T_{1}(x_{i})$ and $\widetilde{x}^{2}_{i} = T_{2}(x_{i})$ where $T_{1}, T_{2}$ are separate augmentation operators sampled from the same family of augmentations $\mathcal{T}$. The Moco objective treats each $(\widetilde{x}^{1}_{i} , \widetilde{x}^{2}_{i})$ and $(\widetilde{x}^{2}_{i}, \widetilde{x}^{1}_{i})$ as a query-positive key pair $x^{query}, x^{key_{pos}}$. Query data points are passed through a query encoder to obtain query features $q = f_{\theta_q}(x^{query})$ and key data points are passed through the key encoder to obtain the key features $k_{pos} = f_{\theta_k}(x^{key_{pos}})$.
The contrastive loss for Moco, $\mathcal{L}_{moco}$ is then given by Eqn. \ref{eqn:Moco}:

\begin{equation}
    \mathcal{L}_{moco} = - log\frac{q\cdot k_{pos}/\tau}{\sum_{i=0}^{K}exp(q\cdot k_{i}/\tau)}
    \label{eqn:Moco}
\end{equation}
where $k_{pos}$ is the positive key for each query $q$, whilst $k_{i}$ are previous outputs of the key encoder which were saved in a queue and $\tau$ refers to a temperature hyperparameter.

The second type of contrastive loss considered in this paper is the Supervised Contrastive (SupCLR) loss \cite{khosla2020supervised}. The SupCLR loss is a generalization of the instance discrimination loss which can deal with an arbitrary number of positives samples to enable contrasting samples in a class-wise manner instead of an instance-wise manner. In this case, all the data points in the same class as the query $x^{query}$ are treated as positive samples $x^{key_{pos}}$ whilst data points in a different class as the query are treated as negative samples $x^{key_{neg}}$. The SupCLR loss  only requires using a single encoder $f_{\theta}$ to obtain features for the queries $q$, positive keys $k_{pos}$ and negative keys $k_{neg}$ where $q = f_{\theta}(x^{query})$,
$k_{pos} = f_{\theta}(x^{key_{pos}})$ and $k_{neg} = f_{\theta}(x^{key_{neg}})$. The SupCLR contrastive loss is then given by Eqn. \ref{eqn:SupCLR}:
\begin{equation}
\mathcal{L}_{SupCLR} = \frac{-1}{|P(q))}\sum_{pos \in P(q)}log\frac{exp(q\cdot k_{pos}/\tau)}{\sum_{i \in A(q)} exp(q\cdot k_{i}/\tau)}
    \label{eqn:SupCLR}
\end{equation}
Here, $P(q)$ is the set of indices of all positives keys of $q$ which are present in the batch, and $|P(q)|$ is its cardinality. $A(q)$ is the set of all indices in the batch excluding $q$ itself, this includes both $k_{pos}$ and $k_{neg}$. Intuitively, the instance discrimination loss learns the features in each instance which makes it different from other instances whilst also being invariant to data augmentation, whilst the SupCLR loss learns the features in common between different instances of the same class.

\section{OOD Experimental Preliminaries} \label{Experimental setup}

In this section, we discuss the experimental setup that we use throughout this work.

\noindent\textbf{Datasets} We study OOD detection for the following ID dataset $(D_{in})$ and OOD dataset $(D_{out})$ pairs.
We use both simple grayscale dataset pairs and more complex RGB dataset pairs for the task. For the grayscale dataset case which uses MNIST, FashionMNIST, and KMNIST, one of the datasets is defined as $D_{in}$ whilst the other datasets are defined as $D_{out}$ \cite{lecun1998mnist,xiao2017fashion,clanuwat2018deep}. For the RGB datasets, we use CIFAR-10, CIFAR100 as well as SVHN and follows the same procedure where one dataset is defined as $D_{in}$ whilst the others are defined as $D_{out}$ \cite{krizhevsky2009learning,netzer2011reading}. 

Similar to Winkens et al, we use the class-wise confusion log probability (CLP) as a measure of the difficulty of an OOD detection task \cite{winkens2020contrastive}. As described in Winkens et al, CLP is based on the probability of a classifier confusing OOD data with ID data. This involves computing an estimate of the expected probability of a test OOD sample $x$ being predicted to be an ID class $k$ by an ensemble of $N$ classifiers as given by:
\begin{equation}
c_k(x) = \frac{1}{N}\sum _{j=1}^Np^{j}(\widehat{y}=k|x)    
\end{equation}
where $p^{j}(\widehat{y}=k|x)$ refers to the probability that the $j^{th}$ classifier assigns $x$ as belonging to class $k$.
The class-wise confusion log probability (CLP) of ID class $k$ for $\mathcal{D}_{test}$ becomes:
\begin{equation}
    CLP_{k}(\mathcal{D}_{test}) = log \left ( \frac{1}{\mathcal{D}_{test}} \sum _{x \in \mathcal{D}_{test}} c_k(x) \right )
\end{equation}
We compute the CLP with $\mathcal{D}_{test}$ being the test samples belonging to the OOD dataset. The distance between the dataset pairs is defined by the min-max bounds on the CLP (lowest CLP value to highest CLP value for the dataset pair). In our work, we use 3 Resnet-18 models to calculate $c_k(x)$. For further details on the Confusion Log probability, see Winkens et al \cite{winkens2020contrastive}, section 4.

\begin{table}[h!]
\centering
\begin{tabular}{|p{5.0cm}|c|}
\hline
\textbf{Datasets} &   \textbf{CLP bounds} \\ \hline
 
ID:MNIST, OOD:FashionMNIST  & -7.26 to -7.15 \\ \hline
ID:MNIST, OOD:KMNIST        & -7.29 to -6.88 \\ \hline
ID:FashionMNIST, OOD:MNIST  & -7.43 to -7.23 \\ \hline
ID:FashionMNIST, OOD:KMNIST & -7.31 to -6.54 \\ \hline
ID:KMNIST, OOD:MNIST        & -6.87 to -6.03 \\ \hline
ID:KMNIST, OOD:FashionMNIST & -7.22 to -6.96 \\ \hline
ID:CIFAR10, OOD:SVHN        & -7.25 to -6.27 \\ \hline
ID:CIFAR10, OOD:CIFAR100    & -4.80 to -3.53 \\ \hline
ID:CIFAR100, OOD:SVHN       & -9.08 to -7.52 \\ \hline
ID:CIFAR100, OOD:CIFAR10    & -7.74 to -4.94 \\ \hline
 
\end{tabular}
\caption{Class-wise Confusion Log Probability (CLP) min-max bounds for the different ID-OOD dataset pairs.}
    \label{table:CLP}
\end{table}
\noindent The min-max bounds of the CLP for the different dataset pairs are given in Table \ref{table:CLP}. For the purpose of categorizing the different dataset pairs, we label pairs that have both values for the bounds of the CLP above 6.5 as a far-OOD dataset pair, below 6.5 as near-OOD dataset pair, and dataset pairs which have bounds that have a value above and below 6.5 as a near \& far-OOD dataset pair. The CLP values were chosen by hand so as to separate the datasets and enable comparison.


\noindent\textbf{Metrics} 
We measure the quality of OOD detection using the established metrics for this task, which are the AUROC, AUPR and the FPR at 95\% TPR \cite{hendrycks2016baseline}.
For the AUPR, we treat OOD examples as the positive class. Unless otherwise stated, metrics reported in this work are obtained based on 8 repeat readings.

\noindent\textbf{Training Setup} Experiments were conducted using 2 Tesla P100-PCIE-16GB with 28 CPUs using Pytorch Version 1.7.1 and Pytorch Lightning Version 1.2.3 \cite{paszke2017automatic,falcon2019pytorch}. We adopt a Resnet-50 as the encoder $f_{\theta}$ for all the different models with a fixed-dimensional output dimensionality of 128-D with the outputs being l2 normalized \cite{he2016deep}. For the case of the classification network, also referred to as the Cross-Entropy (CE) model, the output of the Resnet-50 encoder is followed by an additional fully connected layer with an output dimensionality equal to the number of classes in the ID training data. 

For the case of the Moco model, both the query encoder and the key encoder have the same architecture.  All models are trained for 300 epochs with a batch size of 256, using the SGD optimizer with a learning rate of $3e^{-2}$, optimizer momentum of 0.9 and weight decay of 1$e^{-4}$, queue size of 4096, and an encoder momentum $m$ of 0.999. Furthermore, both the Moco and SupCLR model use a softmax temperature $\tau$ of 0.07 in the loss function.

For data augmentation, we use random crop and resize (with random flip), color distortion (for the RGB datasets), and Gaussian blur.

\section{Instance discrimination is effective for Far-OOD detection}
\label{Point_OOD_results}
\subsection{Hypothesis}
We hypothesized that as instance discrimination requires being able to distinguish each individual sample from one another, the model trained using instance discrimination will output features $z$ which retains a large amount of information about the input $x$. By retaining a large amount of information regarding an input, the Moco model could
achieve high OOD performance. This depends on how semantically similar the ID and OOD points are to one another and what information the inference approach focuses on.

\subsection{Procedure}

To investigate when the instance discrimination loss is effective, we compare the Moco model with the SupCLR and CE model to see the difference when using training labels in the training process. Furthermore, we look at the effect of supervised labels during inference by examining the OOD detection performance of different models across several different dataset pairs using two different inference approaches. The inference approaches used are the Mahalanobis Distance with class-dependent covariance matrices \cite{winkens2020contrastive} which use class labels, as well as the kernel density estimation (KDE) approach which does not use class labels.

\subsection{Results} 

\begin{table*}[h!]
\centering
\begin{tabular}{lllll}

\toprule
ID & OOD & AUROC & AUPR & FPR \\ 
\midrule
&   & \multicolumn{3}{c}{CE/Moco/SupCLR} \\
\cmidrule(lr){2-5} 
CIFAR10 & SVHN         &  0.950 /  0.574 /   \textbf{0.962*} &  0.975 /  0.779 /   \textbf{0.980*} &  \textbf{0.166} /  0.887 /   0.170 \\ 
 & CIFAR100     &  \textbf{0.883} /  0.576 /   0.882 &  0.863 /  0.591 /   \textbf{0.864*} &  \textbf{0.399} /  0.888 /   0.415 \\ 
 & MNIST        &  \textbf{0.939} /  0.462 /   0.900 &  \textbf{0.920} /  0.437 /   0.880 &  \textbf{0.219} /  0.875 /   0.346 \\ 
 & FashionMNIST &  \textbf{0.955} /  0.514 /   0.943 &  \textbf{0.943} /  0.487 /   0.933 &  \textbf{0.162} /  0.887 /   0.232 \\ 
 & KMNIST       &  \textbf{0.941} /  0.472 /   0.935 &  0.911 /  0.444 /   \textbf{0.915*} &  \textbf{0.188} /  0.893 /   0.233 \\ 
  \hline CIFAR100 & SVHN         &  \textbf{0.853} /  0.567 /   0.828 &  \textbf{0.907} /  0.767 /   0.897 &  \textbf{0.415} /  0.879 /   0.482 \\ 
 & CIFAR10      &  0.726 /  0.569 /   \textbf{0.736*} &  0.680 /  0.562 /   \textbf{0.693*} &  0.784 /  0.881 /   \textbf{0.722*} \\ 
 & MNIST        &  \textbf{0.646} /  0.307 /   0.594 &  \textbf{0.620} /  0.378 /   0.564 &  \textbf{0.719} /  0.964 /   0.775 \\ 
 & FashionMNIST &  \textbf{0.851} /  0.522 /   0.841 &  \textbf{0.812} /  0.484 /   0.806 &  \textbf{0.458} /  0.881 /   0.485 \\ 
 & KMNIST       &  0.757 /  0.403 /   \textbf{0.769*} &  0.720 /  0.413 /   \textbf{0.736*} &  \textbf{0.608} /  0.945 /   0.643 \\ 
  \hline MNIST & FashionMNIST &  0.984 /  0.587 /   \textbf{0.992*} &  0.979 /  0.621 /    0.99* &  0.080 /  0.899 /   \textbf{0.034*} \\ 
 & KMNIST       &  0.957 /  0.616 /   \textbf{0.982*} &  0.958 /  0.694 /    0.98* &  0.218 /  0.899 /   \textbf{0.077*} \\ 
  \hline FashionMNIST & MNIST  &  \textbf{0.807} /  0.624 /   0.754 &  \textbf{0.818} /  0.606 /   0.804 &  \textbf{0.614} /  0.821 /   0.820 \\ 
 & KMNIST &  \textbf{0.821} /  0.638 /   0.791 &  \textbf{0.822} /  0.611 /   0.821 &  \textbf{0.576} /  0.771 /   0.738 \\ 
  \hline KMNIST & MNIST        &  \textbf{0.972} /  0.656 /   0.971 &  \textbf{0.970} /  0.712 /   0.966 &  0.108 /  0.822 /   \textbf{0.107*} \\ 
 & FashionMNIST &  \textbf{0.985} /  0.620 /   \textbf{0.985} &  \textbf{0.984} /  0.621 /   0.978 &  \textbf{0.056} /  0.818 /   \textbf{0.056} \\ 
\bottomrule
 
\end{tabular}
\caption{AUROC, AUPR and FPR for different models on different ID-OOD dataset pairs using the KDE approach and * indicates that an approach is better than the CE baseline.}
\label{tab:datasets_comparison_KDE_SupCLR_KDE_CE_and_KDE_Moco}
\end{table*}

\begin{table*}[h!]
\centering
\begin{tabular}{lllll}

\toprule
ID & OOD & AUROC & AUPR & FPR \\ 
\midrule
&   & \multicolumn{3}{c}{CE/Moco/SupCLR} \\
\cmidrule(lr){2-5} 
CIFAR10 & SVHN         &  0.891 /  0.908* /   \textbf{0.950*} &  0.939 /  0.960* /   \textbf{0.973*} &  0.283 /  0.408 /   \textbf{0.187*} \\ 
 & CIFAR100     &  0.875 /  0.784 /   \textbf{0.902*} &  0.851 /  0.778 /   \textbf{0.892*} &  0.418 /  0.657 /   \textbf{0.386*} \\ 
 & MNIST        &  0.931 /  \textbf{0.988*} /   0.951* &  0.906 /  \textbf{0.985*} /   0.935* &  0.214 /  \textbf{0.051*} /   0.176* \\ 
 & FashionMNIST &  0.945 /  0.971* /   \textbf{0.973*} &  0.926 /  \textbf{0.971*} /   0.965* &  0.180 /  0.143* /   \textbf{0.111*} \\ 
 & KMNIST       &  0.929 /  0.952* /   \textbf{0.956*} &  0.892 /  \textbf{0.941*} /   0.933* &  0.195 /  0.174* /   \textbf{0.142*} \\ 
  \hline CIFAR100 & SVHN         &  \textbf{0.858} /  0.828 /   0.818 &  0.900 /  \textbf{0.908*} /   0.870 &  0.420 /  0.565 /   \textbf{0.413*} \\ 
 & CIFAR10      &  0.731 /  0.614 /   \textbf{0.746*} &  0.680 /  0.574 /   \textbf{0.698*} &  0.730 /  0.836 /   \textbf{0.683*} \\ 
 & MNIST        &  \textbf{0.656} /  0.398 /   0.625 &  \textbf{0.623} /  0.467 /   0.604 &  \textbf{0.754} /  0.964 /   0.811 \\ 
 & FashionMNIST &  0.886 /  0.895* /   \textbf{0.902*} &  0.856 /  \textbf{0.880*} /   0.877* &  0.375 /  0.403 /   \textbf{0.319*} \\ 
 & KMNIST       &  0.777 /  0.604 /   \textbf{0.798*} &  0.730 /  0.595 /   \textbf{0.762*} &  \textbf{0.588} /  0.887 /   0.599 \\

  \hline MNIST & FashionMNIST &  0.988 /  \textbf{0.997*} /   0.990* &  0.985 /  \textbf{0.995*} /   0.983 &  0.018 /  \textbf{0.012*} /   0.028 \\ 
 & KMNIST       &  0.984 /  0.972 /   \textbf{0.993*} &  0.982 /  0.962 /   \textbf{0.991*} &  0.061 /  0.108 /   \textbf{0.030*} \\ 
  \hline FashionMNIST & MNIST  &  0.921 /  \textbf{0.980*} /   0.971* &  0.933 /  \textbf{0.980*} /   0.969* &  0.456 /  \textbf{0.105*} /   0.138* \\ 
 & KMNIST &  0.941 /  0.966* /   \textbf{0.972*} &  0.950 /  0.965* /   \textbf{0.969*} &  0.309 /  0.173* /   \textbf{0.138*} \\ 
  \hline KMNIST & MNIST        &  0.973 /  0.957 /   \textbf{0.991*} &  0.970 /  0.954 /   \textbf{0.989*} &  0.089 /  0.208 /   \textbf{0.037*} \\ 
 & FashionMNIST &  0.962 /  0.982* /   \textbf{0.984*} &  0.953 /  \textbf{0.978*} /   0.973* &  0.219 /  0.076* /   \textbf{0.043*} \\ 
\bottomrule
 
\end{tabular}
\caption{AUROC, AUPR and FPR for different models on different ID-OOD dataset pairs using the Mahalanobis Distance inference approach and * indicates that an approach is better than the CE baseline}
\label{tab:datasets_comparison_Mahalanobis_SupCLR_Mahalanobis_CE_and_Mahalanobis_Moco}
\end{table*}

For the case of the KDE approach, Table \ref{tab:datasets_comparison_KDE_SupCLR_KDE_CE_and_KDE_Moco} shows that the Moco model consistently performed worse. This showed that in the absence of any training labels, the instance discrimination training makes it difficult to group ID data together and therefore leads to poor OOD detection performance. Furthermore, the CE model is generally the highest performing model across the metrics, even outperforming SupCLR. This shows that in the absence of labels during the inference process, the CE model is best able to group the data in the feature space such that the OOD test data are farther than the ID test data from the  ID training data points.

For the task of OOD detection using the Mahalanobis Distance inference approach, Table \ref{tab:datasets_comparison_Mahalanobis_SupCLR_Mahalanobis_CE_and_Mahalanobis_Moco} shows that for the grayscale datasets, the Moco model is competitive with the CE model and the SupCLR model achieves the best results. This shows that for the grayscale datasets, the features learned from contrastive learning are able to compete with or outperform the CE model. However, for the case of the RGB datasets, the Moco approach outperforms the CE model on the CIFAR10-SVHN dataset pair and outperforms SupCLR on the CIFAR100-SVHN dataset pair, both of which are categorized as far-OOD. In contrast, for the case of CIFAR10-CIFAR100 (near-OOD) and CIFAR100-CIFAR10 (near \& far-OOD), Moco performs significantly worse than both supervised approaches. 
This suggests that for far-OOD detection tasks, it can be beneficial to use the unsupervised instance discrimination training to perform OOD detection, although, class discriminatory information obtained from supervised training is important for high performance in near-OOD detection settings. Furthermore, it can be seen that using class labels to group the features during the inference process leads to a larger improvement in the contrastive learning models compared to the CE model as shown by a larger improvement in the contrastive models when using the Mahalanobis Distance inference approach rather than KDE. 

\section{OOD samples are classified into classes that learn non-class discriminatory statistics}\label{overall_class_divergence_analysis}

\subsection{Hypothesis}
We hypothesized that the more similar a class distribution is to the distribution of the entire dataset, which we refer to as the overall distribution, the more likely it is that OOD data will be misclassified as that class. The intuition here is that OOD data tends to be misclassified as belonging to the least distinctive class. We believe that the similarity of a class distribution to the overall distribution indicates how much the features of a particular class contain non-class discriminatory information, which indicates how distinctive the class distribution is.

\subsection{Procedure}
To quantify the dissimilarity of the class distribution to the overall distribution, we approximated the class distribution as well as the overall distribution by Gaussian distributions and calculated the KL divergence between the overall distribution and class distribution, $KL \left( Overall \middle\| Class \right)$. Furthermore, the expected Overall-Class KL as shown by Eqn. \ref{eqn:expected_overall_class_kl} was calculated for the contrastive models to see whether there was any difference in this metric between the two approaches.
\begin{equation}
\mathbb{E}\left [ KL \left( Overall \middle\| Class \right) \right ] = \sum _{k}^K KL \left( Overall \middle\| Class_{k} \right)
\label{eqn:expected_overall_class_kl}
\end{equation}
\subsection{Results}
\begin{figure}[h!]
    
    \includegraphics[width=\columnwidth]{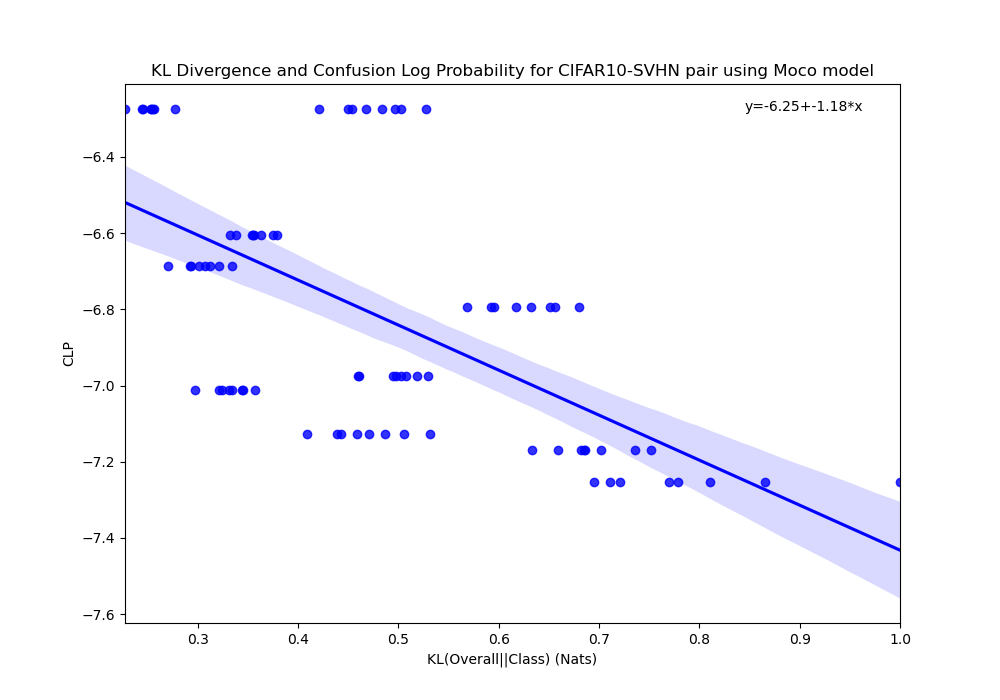}
    \caption{Scatter plot of the normalized KL divergence between the overall distribution and class distribution, $KL \left( Overall \middle\| Class \right)$ (normalized by dividing by the maximum $KL \left( Overall \middle\| Class \right)$), and the class-wise confusion log probability (CLP).}
    \label{fig:KL_CLP}
\end{figure}
From Fig. \ref{fig:KL_CLP}, it can be seen that there is a negative correlation between the KL divergence and the class-wise confusion log probability. This supports our hypothesis that the more dissimilar the class distribution is to the overall distribution, the more likely OOD data points will be misclassified into that ID class. This behavior agrees with the phenomena seen in deep generative models where OOD data points are given high likelihoods by arbitrarily having low compressed lengths and being similar to the non-class discriminatory statistics of the data \cite{serra2019input}. 

\begin{table}[h!]
    \centering
\begin{tabular}{|p{3cm}|p{1.25cm}|p{1.25cm}|}
\hline
\textbf{Datasets} &     \textbf{Moco} &    \textbf{SupCLR} \\ \hline
 
MNIST        &  173.790 &   567.539 \\ \hline
FashionMNIST &  249.147 &  1484.179 \\ \hline
KMNIST       &  122.158 &   468.819 \\ \hline
CIFAR10      &   44.395&   193.645 \\ \hline
CIFAR100     &   96.943 &  1603.609 \\ \hline
\end{tabular}
\caption{Expected Overall-Class KL between the overall distribution and class distribution, $\mathbb{E}\left [ KL \left( Overall \middle\| Class \right) \right ]$, when using different learning approaches.}
    \label{table:KL_div}
\end{table}

From Table \ref{table:KL_div}, it can be seen that the Moco model had expected Overall-Class KL values smaller than the SupCLR model. We believe this explains why the Moco model performs poorly on the near-OOD detection pairs such as CIFAR10-CIFAR100 and CIFAR100-CIFAR10. For the case of Moco where there is a small KL divergence between the overall and class distributions, it indicates that the features from Moco capture general non-class discriminatory information which is common in the ID dataset. This can make near-OOD detection difficult as the OOD dataset is likely to have similar non-class discriminatory statistics in common with the ID dataset. However, in the case of a far-OOD detection situation, the non-class discriminatory statistics are sufficiently different between the two datasets to enable Moco to effectively detect OOD data.

\section{Number of directions of significant variance  }\label{variation_factors_analysis}
\subsection{Hypothesis}
From the OOD detection results (Section \ref{Point_OOD_results}), contrastive learning results in poorer performance in OOD detection than the traditional CE model when using KDE than when using the Mahalanobis Distance approach. We claim that this is, at least in part, due to contrastive learning trained models learn a rich feature space with several different directions of high variance \cite{winkens2020contrastive}. 

In the situation where KDE is used for inference, having a feature space with several directions of variance is likely to result in the OOD data having similar features to the ID data, leading to the misclassification of OOD data as ID. However, in the situation where the Mahalanobis Distance inference approach is used, using class labeled data can help to identify which factors of variation are important for class discrimination which can be used to aid in discriminating the ID data from the OOD data. 

\subsection{Procedure}
 To quantify how rich a feature space is, we investigated the number of factors of variation in the features by following the procedure taken in Section 5 of Roth et al which involves calculating the spectral decay $\rho$ \cite{roth2020revisiting}. This involves computing the singular value decomposition (SVD) of the features of the training data, normalizing the sorted spectrum of singular values, and computing the KL-divergence between a D-dimensional discrete uniform distribution and the sorted spectrum of singular values. Using this metric, lower values of $\rho$ indicate more directions of significant variance.

\subsection{Results}
\begin{figure}[h!]
    \includegraphics[width=\columnwidth]{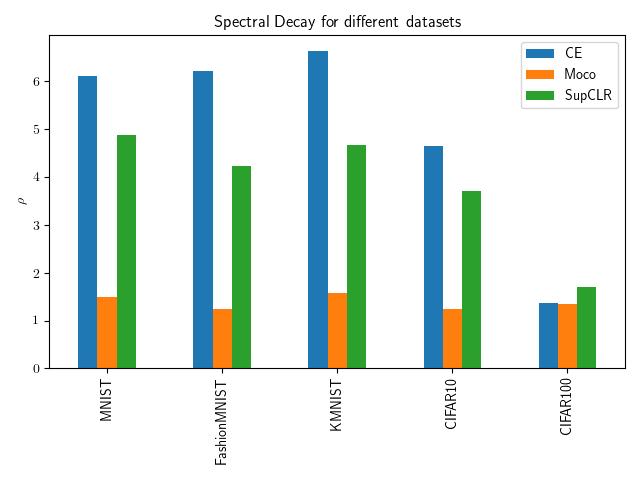}
    \caption{Spectral decay for the different training datasets using different learning approaches.}
    \label{fig:spectral_decay}
\end{figure}
From Fig. \ref{fig:spectral_decay}, it can be seen that both contrastive models generally have lower values of $\rho$ than the CE model, with the exception of the CIFAR100 dataset where the CE model has a lower $\rho$ than the SupCLR approach. As the supervised SupCLR model was able to outperform the supervised CE model on all the datasets in which the SupCLR model had a lower $\rho$ value when using the Mahalanobis Distance, this suggests that having a large number of significant directions of variance is an important property for OOD detection. The Moco model consistently has the lowest $\rho$ which indicates its features have the largest number of factors of variation. This is likely due to the difficult task of instance discrimination requiring learning a large number of factors of variation to perform the task. Whereas, the CE and SupCLR models only require learning features needed to discriminate data between different classes resulting in a smaller number of significant factors of variation.
\section{Related Work}

Recent approaches in OOD detection can be categorized as calibration-based \cite{lee2018simple,liang2017enhancing}, density estimation-based \cite{malinin2018predictive},  and self-supervised learning-based. Self-supervised learning-based approaches can be split into auxiliary pretext task-based \cite{golan2018deep}, contrastive based \cite{winkens2020contrastive}  or approaches that utilize aspects of both auxiliary pretext tasks and contrastive learning \cite{tack2020csi,sohn2020learning}.  The contrastive learning approaches vary significantly and our work also falls in the category of investigating contrastive learning for OOD detection. Some approaches use instance discrimination-based pretraining followed by fine-tuning with a supervised cross-entropy loss \cite{winkens2020contrastive,roy2021does}. Both these papers and ours focus on the instance discrimination contrastive learning task, however, our work focuses on instance discrimination in the absence of fine-tuning with a supervised loss.
Other approaches do not use the pretraining followed by fine-tuning paradigm and instead focus on using a supervised contrastive loss for OOD detection when labels are available \cite{sehwag2021ssd,tack2020csi,cho2021masked}. Our work also examines the performance of the supervised contrastive loss for OOD detection, although  we use the original supervised contrastive loss for the purpose of analyzing the properties present in contrastive learning which makes it effective for OOD detection. Furthermore, our work differs from all the aforementioned work as it is the first to examine the differences between using a supervised contrastive compared to an instance discrimination approach as well as the effect of different inference methods on OOD detection performance.

\section{Conclusion}
In this work, we investigated the effectiveness and limitations of two popular contrastive learning approaches for the problem of OOD detection. In our work, we saw that:
\begin{itemize}
    \item Contrastive learning learns a larger number of factors of variation compared to classification networks. This can be detrimental to OOD detection performance when using KDE (unsupervised inference) but beneficial when using Mahalanobis distance (supervised inference). 
    \item Unsupervised instance discrimination learns a richer feature space with a wide variety of factors of variation compared to supervised contrastive learning as shown by analyzing the spectral decay.
    \item By having a large number of factors of variation, instance discrimination can be more effective than the Cross-Entropy baseline at far-OOD detection.
    \item Despite having a larger number of factors of variation than the supervised contrastive Learning, instance discrimination is ineffective at near-OOD detection.
    \item Instance discrimination is ineffective at near-OOD detection as it leads to a poorer separation between the different classes in the feature space which was shown by having a lower $\mathbb{E}\left [ KL \left( Overall \middle\| Class \right) \right ]$. 
\end{itemize}  

\subsection{Limitations and Future Work}
We would also like to point out some limitations of our study. Firstly, we focus on two different contrastive
learning methods which uses explicit negatives (e.g. MoCo and SupCLR). We chose to focus on these two methods as these methods are quite general contrastive learning approaches where the only significant difference is is the use of supervision. Therefore, comparing the results of the two methods should give an indication of the effect of supervision in OOD detection performance. We believe that other methods based on explicit negatives such as the InfoNCE and Triplet losses would also exhibit similar behaviour \cite{chen2020simple,chopra2005learning}. An avenue of future work could examine the effect of different properties of self-supervised approaches on OOD detection performance. Example of such properties could be whether a self-supervised approach uses explicit negatives or not, as well as the effect of redundancy reduction principle from Barlow Twins \cite{zbontar2021barlow}. 
Furthermore, after identifying that OOD samples tend to be allocated to classes that have a distribution similar to the overall distribution, future work could involve investigating whether it is possible to regularise the contrastive loss to prevent each class from being too similar to the overall distribution. This could enable the instance discrimination contrastive loss to learn a rich feature space whilst also learning class-specific semantic features which can improve near-OOD detection.

\section*{Acknowledgment}
This research was partially funded by an EPRSC PhD studentship as part of the Centre for Doctoral Training in Future Autonomous and Robotic Systems (grant number  EP/L015293/1). RSR is funded by the UKRI Turing AI Fellowship
EP/V024817/1.
\bibliographystyle{plain}
\bibliography{updated_ref}

\end{document}